\documentclass[10pt,twocolumn,letterpaper]{article}

\usepackage{cvpr}              

\usepackage{graphicx}
\usepackage{amsmath}
\usepackage{amssymb}
\usepackage{booktabs}
\usepackage{multirow}
\usepackage{bm}
\usepackage{pifont}

\usepackage[pagebackref,breaklinks,colorlinks]{hyperref}

\newcommand{\cmark}{\ding{51}}%
\newcommand{\xmark}{\ding{55}}%

\usepackage[capitalize]{cleveref}
\crefname{section}{Sec.}{Secs.}
\Crefname{section}{Section}{Sections}
\Crefname{table}{Table}{Tables}
\crefname{table}{Tab.}{Tabs.}

\def\cvprPaperID{2082} 
\def\confName{CVPR}
\def\confYear{2022}

\begin{document}
	
\title{Audio-Driven 3D Facial Animation from In-the-Wild Videos}

\author{
	Liying Lu\textsuperscript{1} \quad Tianke Zhang\textsuperscript{1,2} \quad Yunfei Liu\textsuperscript{1} \quad  Xuangeng Chu\textsuperscript{1,3} \quad Yu Li\textsuperscript{1}\\
	$^1$ Internation Digital Economy Academy \qquad
	$^2$ Tsinghua University\\
	$^3$ The University of Tokyo\\
        {\small \url{https://faw3d.github.io/}}
}

\maketitle
	
\begin{abstract}
Given an arbitrary audio clip, audio-driven 3D facial animation aims to generate lifelike lip motions and facial expressions for a 3D head. Existing methods typically rely on training their models using limited public 3D datasets that contain a restricted number of audio-3D scan pairs. Consequently, their generalization capability remains limited. In this paper, we propose a novel method that leverages in-the-wild 2D talking-head videos to train our 3D facial animation model. The abundance of easily accessible 2D talking-head videos equips our model with a robust generalization capability. By combining these videos with existing 3D face reconstruction methods, our model excels in generating consistent and high-fidelity lip synchronization. Additionally, our model proficiently captures the speaking styles of different individuals, allowing it to generate 3D talking-heads with distinct personal styles. Extensive qualitative and quantitative experimental results demonstrate the superiority of our method. 
\end{abstract}

	
\begin{figure}[t]
	\centering
	\includegraphics[width=1.0\linewidth]{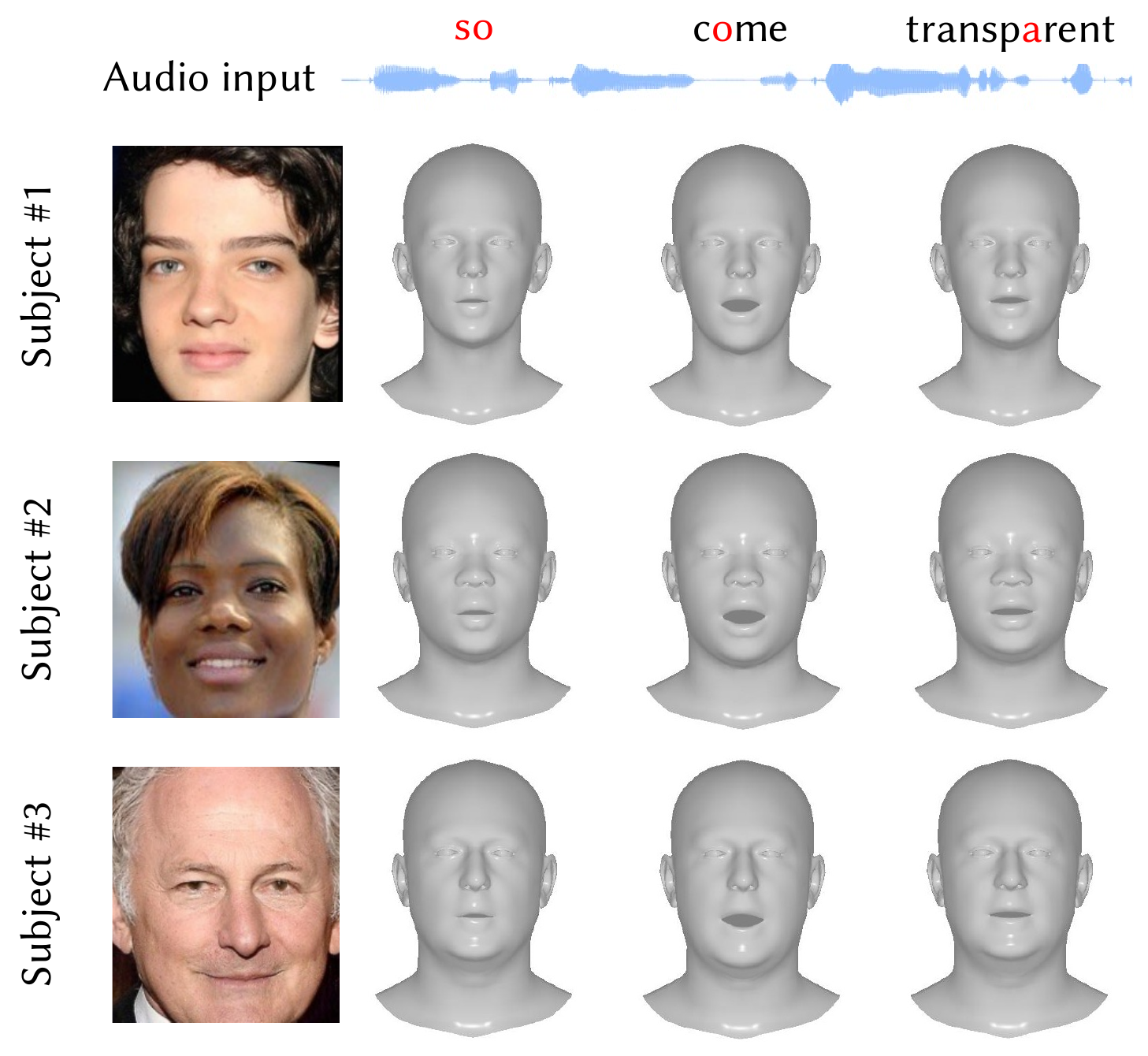}
	\caption{Animation results of our method conditioned on different speech parts with different head shapes.}
	\label{fig:teaser}
\end{figure}

\section{Introduction}

Generating realistic and expressive facial animations is becoming a popular task in computer vision due to its wild real-world applications such as film making, character animation and visual dubbing. Recent advancements in deep neural networks have significantly contributed to the progress of this field. In particular, 2D talking-head video generation aims to synthesize a 2D lip-synced video given an audio clip as well as a reference image or a video of an specific individual, and the generated results should preserve the identity of the individual. They usually leverage existing 2D talking head videos with diverse identities and head poses to train their models. 

On the other, audio-driven 3D facial animation aims to synchronize the lip movements and facial expressions of a 3D head with an arbitrary audio input, thereby creating lifelike and immersive 3D talking-heads effectively. By creating the 3D head model, these methods allow for viewing the generated talking-heads from different camera angles and offer the ability to manipulate illumination, shape, and albedo, resulting in more intricate and vivid results compared to 2D approaches. 

However, existing deep learning-based methods for audio-driven 3D facial animation typically rely on limited public 3D datasets, which are labour-intensive to collect and only contain a restricted number of audio-3D scan pairs. Unfortunately, the scarcity of training data often limits the generalization capability of these models.
To address this challenge, we in this paper propose a novel method that harnesses the abundance of in-the-wild 2D talking-head videos to train our 3D facial animation model. These readily available 2D videos provide a vast and diverse source of facial motion data, equipping our model with a robust generalization capability. By combining these videos with state-of-the-art 3D face reconstruction methods, we construct a dataset called HDTF-3D that encompasses a wide range of 3D facial motion information. As a result, our model, trained on this dataset, excels in generating consistent and high-fidelity lip synchronization, while also proficiently captures the identities and speaking styles of different individuals.
This enables our model to produce identity-specific 3D talking-head videos with remarkable realism.

Specifically, we propose a model that takes an audio clip, a reference image and a style code as input. The reference image is utilized to extract the identity-related information of the speaking person, such as the head shape and albedo. 
Meanwhile, the style code represents the desired speaking style for the output video, including mouth opening and closing amplitudes, as well as expression amplitudes. By processing these inputs, our model generates a 3D identity-specific talking-heads video with high realism and low lip-sync errors. In addition, we further andow our model with the capability of emotion manipulation. By speficying an expression style, such as "\textit{angry}", "\textit{happy}" or "\textit{sad}", our model could produce lip-synced videos with the corresponding expression style. Extensive qualitative and quantitative experimental results demonstrate the superiority of our method, and the contributions of our work are summarized as follows:
\begin{itemize}
  \item \textbf{Utilizing In-The-Wild Videos for Learning} We propose to learn the audio-driven 3d facial animation model from in-the-wild videos that could provide a vast and diverse source of facial motion data. We construct a dataset called HDTF-3D that encompasses a wide range of 3D facial motion information.
  \item \textbf{Strong Generalization Capability} By training our model on the proposed HDTF-3D dataset, we have endowed it with robust generalization capabilities. As depicted in Fig.~\ref{fig:teaser}, our model excels at synthesizing accurate and realistic mouth shapes, even when dealing with subjects who have varying head shapes. 
  \item \textbf{Emotion Manipulation} Our model is equipped with the emotion manipulation capability. By specifying a desired expression style, our model can generate 3D talking-head videos with the corresponding expression. 
\end{itemize}

\section{Related Work}

\noindent\textbf{Audio-Driven 2D Talking-Head Video Generation}
The field of talking-head video generation has garnered considerable attention and has been extensively studied over the past years. The goal of talking-head video generation is to create lifelike videos in which virtual characters or avatars convincingly simulate speaking by accurately synchronizing their lip movements with the accompanying audio. Recently, deep learning-based methods have achieved superior results. Some works~\cite{suwajanakorn2017synthesizing,thies2020neural,guo2021ad,ji2021audio,li2021write,lu2021live,song2022everybody,liu2022semantic} are designed to generate high-quality talking-head videos for specific individuals. For example, Suwajanakorn~\etal~\cite{suwajanakorn2017synthesizing} introduce a compositing approach to synthesize photo-realistic talking-head videos of Barack Obama. Justus~\etal~\cite{thies2020neural} propose to extract a representation of person-specific talking styles from a short video clip and generate the results that preserve the talking style.AD-NeRF~\cite{guo2021ad} and SSP-NeRF~\cite{liu2022semantic} propose to utilize the neural radiance fields~(NeRF) to synthesize vivid results.

On the other hand, some works~\cite{chung2017you,chen2019hierarchical,song2018talking,zhou2019talking,wav2lip,zhou2021pose,zhou2020makelttalk,liang2022expressive,sun2022masked} focus on developing frameworks that are speaker-agnostic. They can address all identities by a single model, by only taking a reference image or a short reference video. Speech2Vid~\cite{chung2017you} is the very first end-to-end model that could synthesize a video of the target face lip synched with the given audio. Wav2Lip~\cite{wav2lip} propose to train an expert lip-sync discriminator to improve the Lip-sync accuracy. Sun~\etal~\cite{sun2022masked} propose to leverage cross-frame and cross-modal context information to predict the masked mouth region and design a Transformer to synthesize more accurate results.

\vspace{5pt}
\noindent\textbf{Audio-Driven 3D Facial Animation}
Different from the 2D talking-head generation, audio-driven 3D facial animation aims to generate 3D face meshes driven by the input audio. Earlier procedural methods~\cite{edwards2016jali,massaro201212,taylor2012dynamic,xu2013practical,kalberer2001face} focus on developing different explicit rules to animate a predefined facial rig. However, these methods require a lot of manual effort. With the development of deep learning, many data-driven methods~\cite{liu2015video,cudeiro2019capture,hussen2020modality,karras2017audio,pham2018end,richard2021meshtalk,faceformer,codetalker} have been proposed and achieve superior results. They generally first extract the representation of the given audio and then feed it into a CNN model to obtain the target 3D mesh. VOCA~\cite{cudeiro2019capture} takes as input a subject-specific 3D template and an audio signal and processes them through an encoder-decoder network, producing the vertex displacements from the template. MeshTalk~\cite{richard2021meshtalk} propose to learn a categorical latent space of facial animation that disentangles audio-correlated and audio-uncorrelated information. FaceFormer~\cite{faceformer} introduces a Transformer architecture and predicts face motions in an autoregressive way. CodeTalker~\cite{codetalker} proposes a discrete motion prior-based model. It first learns a codebook together with a decoder that stores the realistic facial motion priors in a self-reconstruction way, then uses the learned modules to map the input audio to the facial motions. These data-driven methods typically rely on 3D datasets with audio-3D scan pairs. However, the number of data in existing 3D datasets is limited.

\vspace{5pt}
\noindent\textbf{Talking-Head Video Datasets}
They are bunches of 2D talking-head video datasets available for learning the 2D talking-head generation models. For example, VoxCeleb2~\cite{chung2018voxceleb2} is one of the widely used audio-visual datasets in the area of talking-head generation, which consists of over a million utterances from over 6k speakers. Another popular dataset LRS2~\cite{son2017lip} contains mainly news and talk shows from BBC programs. LRS3-TED~\cite{afouras2018lrs3} includes face tracks from over 400 hours of TED and TEDx videos, along with the corresponding subtitles and word alignment boundaries. TalkingHead-1KH consists of 500k video clips from YouTube, of which about 80k are greater than $512 \times 512$ resolution.

Different from 2D datasets, the number of 3D datasets~\cite{cheng20184dfab,fanelli20103,cudeiro2019capture} are very limited due to the labour-intensive collection process. Fanelli~\etal collected the BIWI dataset~\cite{fanelli20103} by employing a real-time 3D scanner and a studio condenser microphone. BIWI contains 40 short English sentences uttered by 14 subjects, including 8 females and 6 males. The total number of the recorded sequences is 1109, with 4.67 s long on average. Cudeiro~\etal collected the VOCASET~\cite{cudeiro2019capture} dataset with 4D face scans and speech. VOCASET has 480 sequences uttered by 12 subjects, each is about 3-4 seconds. The restricted number of training data could result in suboptimal generalization capability. Therefore, in this paper, we propose to utilize the large number of readily available 2D videos to train the 3D facial animation model.


\begin{figure*}[t]
  \centering
  \includegraphics[width=1.0\textwidth]{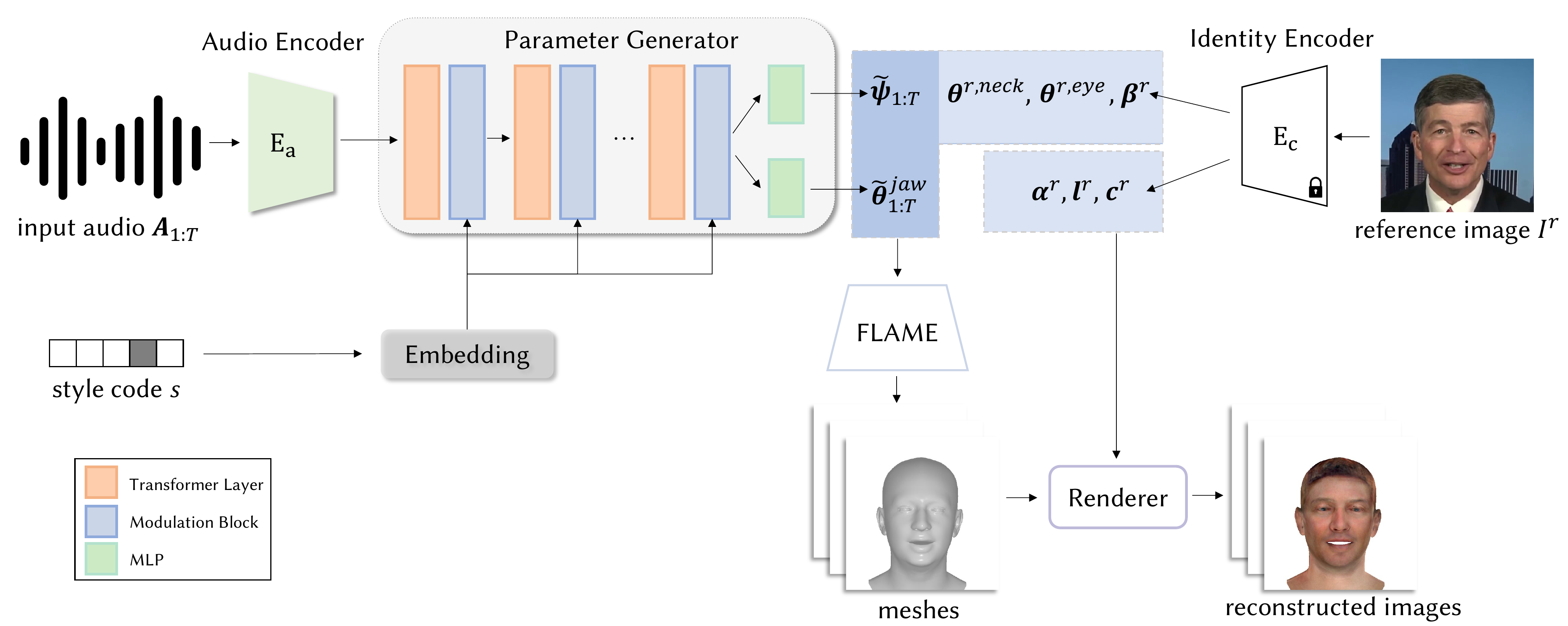}
  \caption{The overview of our framework. $E_a$ and $E_c$ denote the audio encoder and the identity encoder, respectively.}
  \label{fig:framework}
\end{figure*}

\section{Method}

\subsection{HDTF-3D Dataset Construction from In-The-Wild Videos}
\label{sec:data_cons}

We first introduce how to construct our HDTF-3D dataset for 3D facial animation from in-the-wild videos. As mentioned, the number of public 3D datasets that contain audio-3D scan pairs are restricted. However, there are bunches of in-the-wild 2D talking-head videos available for utilization. If we could utilize these 2D videos for learning the 3D facial animation model, the performance and generalization capability of the trained model can be significantly improved. To this end, we propose to leverage the existing 3D face reconstruction method EMOCA-v2~\cite{emoca} as our information extraction tool. 

\vspace{5pt}
\noindent\textbf{EMOCA-v2}
Specifically, EMOCA-v2 is the current state-of-the-art monocular 3D facial reconstruction model that could regress 3D faces from monocular images with fine facial geometric details and high-fidelity expressions. It uses FLAME~\cite{FLAME} as its face model. FLAME models 3D heads with parameters for identity shape $\bm{\beta} \in \mathbb{R}^{\left | \bm{\beta} \right |}$, expression $\bm{\psi} \in \mathbb{R}^{\left | \bm{\psi} \right |}$, and poses $\bm{\theta} \in \mathbb{R}^{3k3k}$~(with $k = 4$ joints for neck, jaw and eyeballs) and outputs a mesh with $n_v = 5023$ vertices:
\begin{align}
  M(\bm{\beta}, \bm{\theta}, \bm{\psi}) \rightarrow (\bm{V}, \bm{F}) \;,
\end{align}
where $\bm{V} \in \mathbb{R}^{n_v \times 3}$ and $\bm{F} \in \mathbb{R}^{n_f \times 3}$~($n_f$ = 9976) are the output vertices and faces.

Based on FLAME, EMOCA-v2 constructs three encoders: a coarse encoder $E_c$, a detail encoder $E_d$ and an expression encoder $E_e$ to extract necessary information from a given image, which, apart from $\bm{\beta}$, $\bm{\psi}$ and $\bm{\theta}$, also includes the albedo $\bm{\alpha} \in \mathbb{R}^{\left | \bm{\alpha} \right |}$, the Spherical Harmonics~(SH)~\cite{ramamoorthi2001efficient} lighting $\bm{l} \in \mathbb{R}^{27}$, and camera $\bm{c} \in \mathbb{R}^{3}$. The extracted information is then fed into the FLAME model and a detail decoder to produce the 3D mesh. In this paper, we leverage EMOCA-v2's pretrained $E_c$ and $E_e$ to produce our needed parameters:
\begin{align}
  E_c(I) &\rightarrow (\bm{\beta}, \bm{\theta}, \bm{\psi}, \bm{\alpha}, \bm{l}, \bm{c}) \label{eq:Ec} \;, \\
  E_e(I) &\rightarrow \bm{\psi}^e \;,
\end{align}
where the expression code $\bm{\psi}$ predicted by the coarse encoder is discarded and the expression code $\bm{\psi}^e$ predicted from the expression encoder is used to produce the final 3D mesh. After that, we can use a differential renderer $R$ to render the reconstructed 2D image $I^{Re}$:
\begin{align}
  R(M(\bm{\beta}, \bm{\theta}, \bm{\psi}^e), \bm{\alpha}, \bm{l}, \bm{c}) \rightarrow I^{Re} \label{eq:render} \;,
\end{align}

Specifically, We use HDTF dataset~\cite{zhang2021flow} as our 2D video source. HDTF is a large in-the-wild high-resolution audio-visual dataset, consisting of about 362 different videos for 15.8 hours, and the resolution of the origin video is 720P or 1080P. We construct a tracker based on EMOCA-v2 which firstly tracks the expressions, the poses and other parameters of the target head in a given video from HDTF, then optimizes the camera poses based on the 2D landmarks obtained from MediaPipe~\cite{lugaresi2019mediapipe}. To smooth the output parameters of each frame, we post-process them by adopting the Kalman Filter, producing a series of more stable and consistent parameters. As a result, we obtain 133 pairs of 3D facial parameters for 3D facial animation learning.

\subsection{Audio-Driven 3D Facial Animation Network}

Audio-driven 3D facial animation aims to synthesize a lip-synced 3D talking-head video from a given audio clip. Similar to previous methods~\cite{faceformer,codetalker}, we formulate it as a sequence-to-sequence problem. Suppose we have a sequence of the ground-truth 3D facial parameters $\bm{X}_{1:T} = (\bm{x}_1, ..., \bm{x}_T)$, where $T$ is the number of frames and $\bm{x}_t$ denotes the 3D facial parameters $(\bm{\beta}_t, \bm{\theta}_t, \bm{\psi}^e_t, \bm{\alpha}_t, \bm{l}_t, \bm{c}_t)$ that corresponds to the $t$-th frame. Then given an audio sequence $\bm{A}_{1:T} = (\bm{a}_1, ..., \bm{a}_T)$, where $\bm{a}_t$ is the audio window that is aligned with the $t$-th frame, as well as a reference image $I^r$ for providing the identity information, our goal is to sequentially generate the 3D facial parameters $\bm{X}_{1:T}^{'} = (\bm{x}_1^{'}, ..., \bm{x}_T^{'})$ that should be as close as the ground-truth sequence. As a result, the synthesized 3D talking-head video will not only be lip-synced with the given audio but also preserve the identity of the reference image.

Our overall framework is illustrated in Fig.~\ref{fig:framework}. It comprises three primary modules: the audio encoder, the identity encoder, and the parameter generator. In the subsequent sections, we provide a comprehensive explanation of each of these modules.

\vspace{5pt}
\noindent\textbf{Audio Encoder}
Our audio encoder follows a similar design to that of Wav2Lip~\cite{wav2lip}. We first pre-process the audios to 16kHz, then convert them to mel-spectrograms with FFT window size 800 and hop length 200. For each frame, we extract a window of 0.2 seconds of mel-spectrogram, centered around the frame's time-step. This process yields a mel-spectrogram feature $\bm{a}_t \in \mathbb{R}^{16 \times 80}$. The audio encoder is composed of fully convolutional layers and residual blocks, which process the mel-spectrogram feature and generate an audio feature $\bm{a}_t^{'} \in \mathbb{R}^{512}$.

\vspace{5pt}
\noindent\textbf{Identity Encoder}
Given a reference image $I^r$, we want the synthesized talking-head video to preserve the identity of $I^r$. This design enables our framework to generate videos with different identities by just changing the input reference image. We utilize the pretrained $E_c$ of EMOCA-v2 as our identity encoder and extract the identity shape, the albedo and the illumination $(\bm{\beta}^r, \bm{\alpha}^r, \bm{l}^r)$ as Eq.~\ref{eq:Ec}. The neck pose $\bm{\theta}^{r, neck}$ and the eyeball pose $\bm{\theta}^{r, eye}$ are also saved for further use.

\vspace{5pt}
\noindent\textbf{Parameter Generator}
As transformers have demonstrated their superiority in modelling long-range dependencies, we adopt transformer layers as the main backbone of our parameter generator. As shown in Fig.~\ref{fig:framework}, it takes the audio feature $\bm{a}_t^{'}$ and a style code $\bm{s}$ as input. The style code is a one-hot vector and is used to control the speaking style of the output video. It is first encoded into a style feature $\bm{s}_{emb}$ by an embedding layer, and then is fed into the transformer layers together with the audio feature. 

Specifically, the audio feature $\bm{a}_t^{'}$ is first mapped through a linear layer and then added with the positional embedding that is used for indicating the temporal order information, producing the processed audio feature $\hat{\bm{a}}_t$. After that, several temporal self-attention layers~(TSA) are used to model the correspondences among the audio features. TSA first produces the key, query, value features $\bm{Q}$, $\bm{K}$ and $\bm{V}\in \mathbb{R}^{(T, d)}$ as:
\begin{align}
  \bm{Q}=\hat{\bm{A}}\bm{W}_Q,~~ \bm{K}=\hat{\bm{A}}\bm{W}_{K},~~ \bm{V}=\hat{\bm{A}}\bm{W}_{V} \;, 
  \label{eq:qkv}
\end{align}
where $\hat{\bm{A}} \in \mathbb{R}^{(T,d)}$ is the audio feature sequence. $\bm{W}_Q$, $\bm{W}_{K}$, and $\bm{W}_{V}$ are projection matrices. Then the self-attention is computed as:
\begin{align}
	Attn(\bm{Q},\bm{K},\bm{V})=Softmax(\frac{\bm{Q}\bm{K}^{T}}{\sqrt{d}})\bm{V} \;.
	\label{eq:attn}
\end{align}
The final feature $\tilde{\bm{A}}$ is obtained by feeding the attention result to a linear layer.

We then use a modulation block to further incorporate the style feature into the generation process. The modulation block first maps the style feature $\bm{s}_{emb}$ into two affine transformation parameters $\bm{\gamma}$ and $\bm{\delta}$ by an multilayer perceptron~(MLP), Then the transformation is carried out by scaling and shifting the audio feature, formulated as:
\begin{align}
	\bm{\gamma}, \bm{\delta} = MLP(\bm{s}_{emb}) \;, \\
  \bm{B} = \bm{\gamma} \cdot \tilde{\bm{A}} + \bm{\delta} \;,
\end{align}
where $\cdot$ denotes Hadamard product and $\bm{B}$ is the resulted feature. 
At last, $\bm{B}$ is used to produce the expression sequence ($\tilde{\bm{\psi}}_1, ..., \tilde{\bm{\psi}}_T$) and the jaw pose sequence ($\tilde{\bm{\theta}}^{jaw}_{1}, ..., \tilde{\bm{\theta}}^{jaw}_{T}$). Since the value range of the expression and jaw pose parameters are different, to obtain rational results, we use two different MLPs to output these two sequences.

\subsection{Training Objectives}
\label{sec:loss}

During training, we freeze the model parameters of the identity encoder and only train the audio encoder and the parameter generator. 
To ensure the accuracy and realism of the synthesized result, we adopt several different loss functions.

\vspace{5pt}
\noindent\textbf{Regression Loss}
Our model is trained in a regressive way. The regression loss computes the L2 difference between the output parameters and the ground-truth parameters as:
\begin{align}
  \mathcal{L}_{reg} = \sum_{t=1}^{T} \left ( (\| \tilde{\bm{\psi}}_t - \bm{\psi}_t \|_{2}) + (\| \tilde{\bm{\theta}}^{jaw}_{t} - \bm{\theta}^{jaw}_{t} \|_{2}) \right ) \;,
\end{align}

\vspace{5pt}
\noindent\textbf{Mouth Closure Loss}
To emphasize the lip-synchronization accuracy, we re-project the generated 3D mesh back to the 2D image and extract the 2D landmarks related to the upper and lower lip, which are compared with the ground-truth 2D lip landmarks as:
\begin{align}
  \mathcal{L}_{mc} = \sum_{t=1}^{T}  \left ( (\| \tilde{\bm{k}}_{x,t} - \bm{k}_{x,t} \|_{1}) + (\| \tilde{\bm{k}}_{y,t} - \bm{k}_{y,t} \|_{1}) \right ) \;,
\end{align}
where $\bm{k}$ includes all 2D landmarks of the upper and lower lip. $\bm{k}_{x,t}$ denotes the x coordinate value of the of the $t$-th frame's ground-truth landmarks.

\vspace{5pt}
\noindent\textbf{Photometric Loss}
The Photometric loss computes the pixel difference between the ground-truth image $I_t$ and the rendered 2D image $I_t^{Re}$ as:
\begin{align}
	\mathcal{L}_{pho} = \sum_{t=1}^{T} \| \bm{M}_t \cdot (I_t - I_t^{Re}) \|_{1} \;,
\end{align}
where $I_t^{Re}$ is rendered as Eq.~\ref{eq:render}. $\bm{M}_t$ denotes the rendered mask of the output face shape, where each pixel within the face skin region is assigned a value of 1, while pixels outside the face skin region are set to 0.

\vspace{5pt}
\noindent\textbf{Emotion Consistency Loss}
Following \cite{emoca}, we also adopt the emotion consistency loss to improve the quality of the emotional content of the output result:
\begin{align}
	\mathcal{L}_{emo} = \sum_{t=1}^{T} \| ENet(I_t) - ENet(I_t^{Re}) \|_{2} \;,
\end{align}
where $ENet$ is a pretrained emotion recognition network with the prediction heads dicarded.

\vspace{5pt}
\noindent\textbf{Overall Objective}
Our full objective is defined as:
\begin{align}
  \mathcal{L} = {\lambda}_{reg}\mathcal{L}_{reg} + {\lambda}_{mc}\mathcal{L}_{mc} + {\lambda}_{pho}\mathcal{L}_{pho} + {\lambda}_{emo}\mathcal{L}_{emo} \label{eq:objective} \;,
\end{align}
where ${\lambda}_{reg}$, ${\lambda}_{mc}$, ${\lambda}_{pho}$ and ${\lambda}_{emo}$ are the weighting factors.

\subsection{Inference}

During inference, aside from the given audio sequence, the reference image and the style code, our model should also be provided with the camera parameters for modeling the head pose, which could be extracted from an external talking-head video. Then the final synthesis process is formulated as:
\begin{align}
  M(\bm{\beta}^{r}, \tilde{\bm{\theta}}, \bm{\psi}) \rightarrow (\bm{V}_t, \bm{F}_t) \;,
\end{align}
where $\tilde{\bm{\theta}} = (\bm{\theta}^{r, neck}, \bm{\theta}^{r, eye}, \tilde{\bm{\theta}}^{jaw}_{t})$.

\begin{figure}[t]
  \centering
  \includegraphics[width=1.0\linewidth]{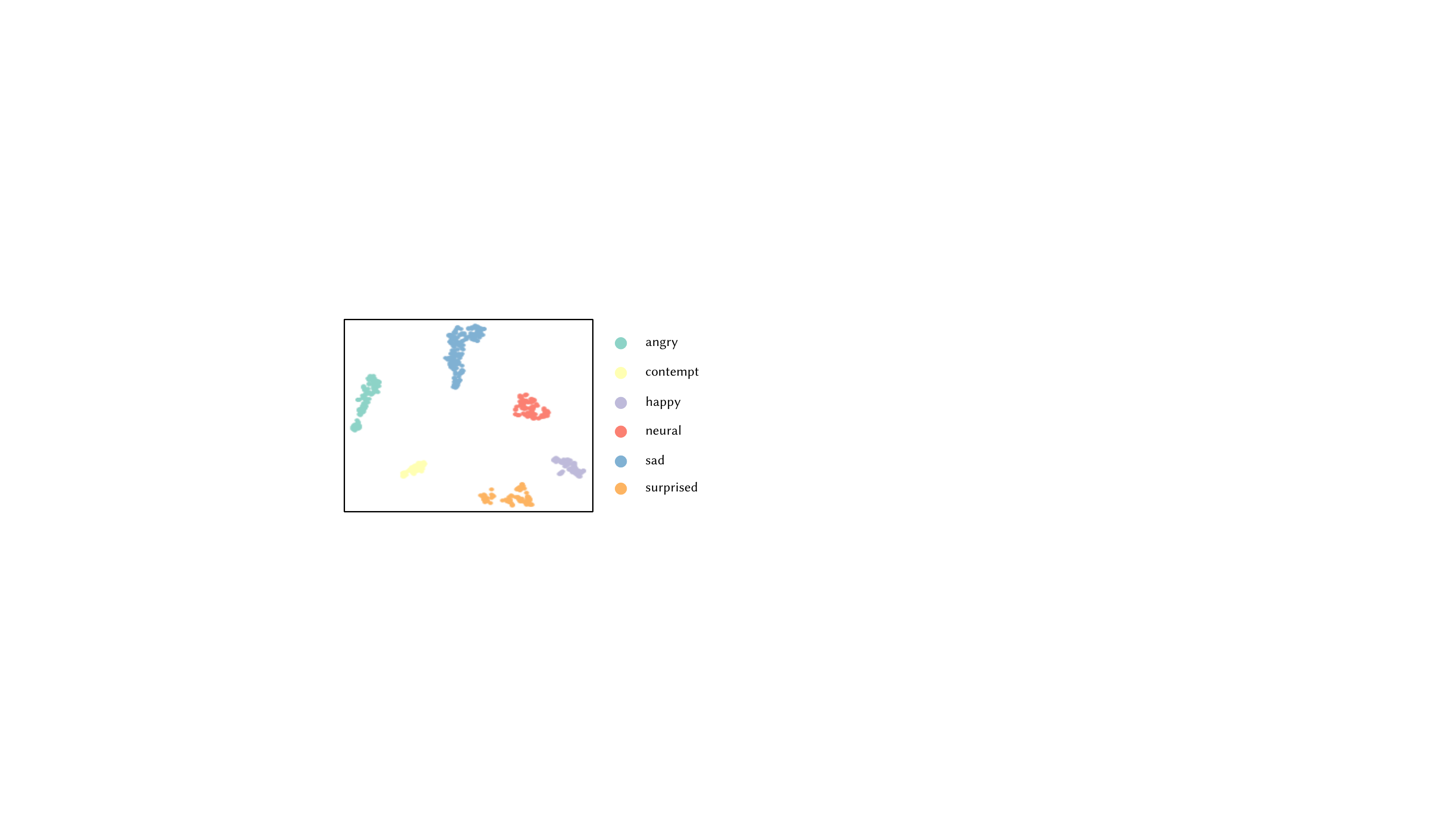}
  \caption{Visualization of expression codes of different emotions. It can be observed that different emotions are clustered in different regions.}
  \label{fig:tsne}
\end{figure}

\subsection{Emotion Manipulation}
\label{sec:manipulation}

Emotion manipulation is a desired function of talking-head generation. It is appealing that given a single audio clip, we can synthesize videos with different emotions. We endow our model with the emotion manipulation capability by analyzing the emotion information in the MEAD dataset~\cite{mead}. MEAD is a talking-head video corpus featuring 60 actors talking with different emotions at three different intensity levels (except for neutral). We utilize videos from MEAD and extract the corresponding emotion template for five emotions, including \textit{angry, contempt, happy, sad, surprise}. 

Specifically, we take the videos with these five emotions at the highest intensity level as our analyzation source. Taking the "angry" emotion as the example, for a video labeled with "angry", we first apply the emotion recognition network on each frame and get the prediction result. If the predicted result is exactly "angry", then this frame is labeled as a valid frame, otherwise labeled as invalid. After predicting the emotion for every frame, we can get all the valid frames of the given video. Then we use the pretrained $E_c$ of EMOCA-v2 to extract the expression code of the valid frames and average them to get our "angry" template $\bm{\psi}^{angry}_{temp}$. 

By repeating the same process, we can get five templates for our desired five emotions. We also use t-SNE~\cite{van2008visualizing} to visualize the expression codes correspond to different emotions, including the neural emotion. As shown in Fig.~\ref{fig:tsne}, different emotions are clustered in different regions, validating the rationality of our method.

After that, by adding the emotion template to the generated expression code, the emotion of the output talking head will be changed accordingly:
\begin{align}
	\tilde{\bm{\psi}}_t^{angry} = \tilde{\bm{\psi}}_t + \bm{\psi}^{\textit{e}}_{temp} \;,
\end{align}
where the superscript $\textit{e}$ = ["angry", "contempt", "happy", "sad", "surprise"].

However, since the expression code is entangled with mouth closing/opening, simply adding the template will result in the mouth closure change and the output video will have inaccurate lip synchronization. Therefore, we further analyze each dimension of the expression code. We found that the $1^{st}$ and the $4^{th}$ dimension of the expression code is related to mouth closing/opening, thus whiling adding the emotion template to the expression code, we will mask out the $1^{st}$ and the $4^{th}$ dimension of the template. In addition, to further control the emotion level, we adopt a weighting scheme. The final emotion manipulation process is formulated as:
\begin{align}
	\tilde{\bm{\psi}}_t^{angry} = \tilde{\bm{\psi}}_t + \bm{w} \cdot \bm{\psi}^{e}_{temp} \;,
\end{align}
where $\bm{w}$ is the weighting vector, whose $1^{st}$ and the $4^{th}$ elements are set to $0$, and the values of other elements are set to be the same, ranging from $[0, 1]$ according to our experiment result. The weighting values determine the intensity level of the emotions, where larger values correspond to higher intensities.


\section{Experiments}

\begin{table}[t]
  \centering
  \scalebox{0.92}{\begin{tabular}{c|cccc}
  \hline Methods & LDE $\downarrow$ & LVE $\downarrow$ & EDE $\downarrow$ & EVE $\downarrow$  \\
  \hline 
  FaceFormer~\cite{faceformer} & 0.0014 & 0.0013 & 0.0030 & 0.0006  \\
  CodeTalker~\cite{codetalker} & 0.0015 & 0.0013 & 0.0032 & 0.0006  \\
  Ours & \textbf{0.0009} & \textbf{0.0011} & \textbf{0.0014} & \textbf{0.0005}  \\
  \hline
  \end{tabular}}
  \caption{Quantitative comparison on the HDTF-3D dataset. \textbf{Bold} indicates the best performance.}
  \label{table:quantitative}
\end{table}

\begin{table}[t]
  \centering
  \scalebox{0.92}{\begin{tabular}{c|cccc}
  \hline Methods & LDE $\downarrow$ & LVE $\downarrow$ & EDE $\downarrow$ & EVE $\downarrow$  \\
  \hline 
  FaceFormer~\cite{faceformer} & 0.0021 & 0.0014 & 0.0037 & 0.0007  \\
  CodeTalker~\cite{codetalker} & 0.0021 & 0.0014 & 0.0036 & 0.0007  \\
  Ours & \textbf{0.0017} & \textbf{0.0014} & \textbf{0.0023} & 0.0007  \\
  \hline
  \end{tabular}}
  \caption{Quantitative comparison on the TalkingHead-1KH dataset. \textbf{Bold} indicates the best performance.}
  \label{table:quantitative_th}
\end{table}

\begin{figure}[t]
  \centering
  \includegraphics[width=1.0\linewidth]{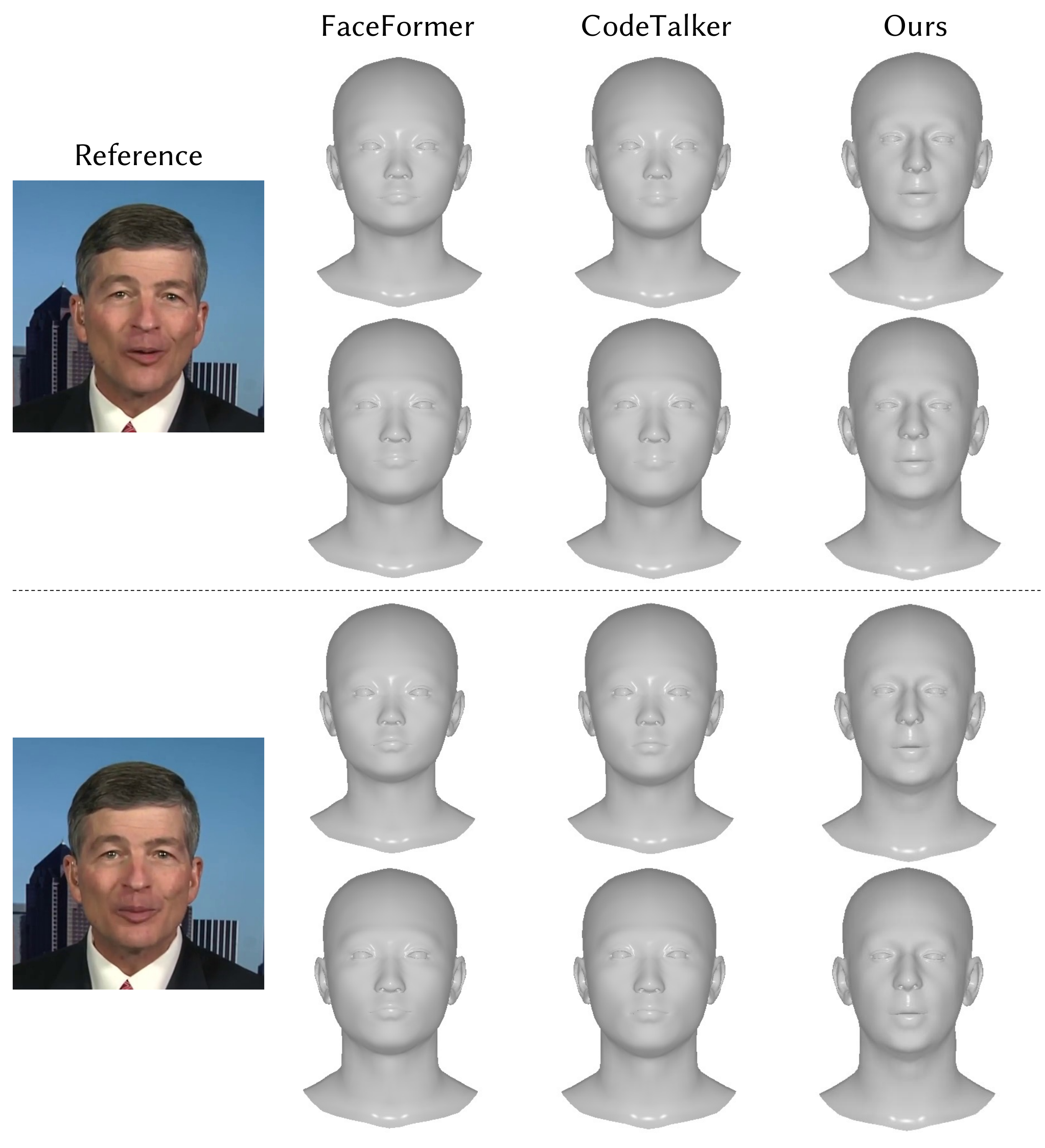}
  \caption{Testing models on head shapes that are unseen during training. Two cases are given here. In each case, the first and the second row corresponds to two different head shapes.}
  \label{fig:change_shape}
\end{figure}

\subsection{Implementation Details}

\noindent\textbf{Dataset}
As mentioned in Sec.~\ref{sec:data_cons}, we construct our HDTF-3D dataset based on the HDTF dataset. Then our HDTF-3D dataset is split into the training set and the testing set, which contains 124 videos and 9 videos respectively. 
In addition, to further validate the generalization capability of our method, we also use the TalkingHead-1KH dataset~\cite{wang2021facevid2vid}. It consists of 500k talking-head video clips from YouTube, from which we randomly select 25 videos for evaluation, their time durations are  24-42s. 

\vspace{5pt}
\noindent\textbf{Implementation Details}
We train our model in two stages. In the first stage, we jointly train the audio encoder and the parameter generator only with the regression loss, and the batch size is set to 128. After xx iterations, they are then jointly trained with the overall objective as Eq.~\ref{eq:objective}, where we set ${\lambda}_{reg} = 1$, ${\lambda}_{mc} = 0.1$, ${\lambda}_{pho} = 5$ and ${\lambda}_{emo} = 1$. The batch size of the second stage is 16. In both stages, we use Adam~\cite{kingma2014adam} as the optimizer with a learning rate of $1e-4$. The temporal length $T$ is set to 12 during both training and testing. The overall framework is implemented in PyTorch~\cite{paszke2019pytorch}, and we use the differentiable rasterizer from Pytorch3D~\cite{ravi2020pytorch3d} for rendering.

\subsection{Quantitative Evaluation}

\noindent\textbf{Lip Distance Error}
To evaluate the quality of lip movements, we define the lip distance error~(LDE), which is measured as the L2 error between the predicted lip vertices $\tilde{\bm{v}}_t$ and the ground truth lip vertices $\bm{v}_t$ for each frame. Here we select the lip vertices that correspond to the inner upper lip and the inner lower lip of the MediaPipe landmark to calculate the error. The final evaluation result is computed by taking the average of the error over all frames:
\begin{align}
	LDE = \frac{1}{T} \sum_{t=1}^{T} \| \tilde{\bm{v}}_t - \bm{v}_t \|_{2} \;,
\end{align}

\vspace{5pt}
\noindent\textbf{Lip Velocity Error}
We further define the lip velocity error~(LVE), which is used for measuring lip motion accuracy. We compare the lip motions of the predicted lip vertices with the ground truth as:
\begin{align}
	LVE = \frac{1}{T-1} \| (\tilde{\bm{V}}_{[2:T]} - \tilde{\bm{V}}_{[1:T-1]}) - (\bm{V}_{[2:T]} - \bm{V}_{[1:T-1]}) \|_{2} \;,
\end{align}
where $\tilde{\bm{V}} = (\tilde{\bm{V}}_{1}, ..., \tilde{\bm{V}}_{T})$ is the lip vertice sequence.

\vspace{5pt}
\noindent\textbf{Expression Distance Error}
Similar to the lip distance error, we compute the expression distance error~(EDE) as the L2 error of all vertices for each frame and take the average over all frames.

\vspace{5pt}
\noindent\textbf{Expression Velocity Error}
Similar to the lip velocity error, the expression velocity error~(EVE) is calculated as the L2 error of the motions of all vertices.

We conducted a comparative analysis of our method against two prominent state-of-the-art 3D facial animation methods, namely FaceFormer and CodeTalker. The quantitative results on the HDTF-3D dataset are summarized in Table.~\ref{table:quantitative}, where it is obvious that our method outperforms both FaceFormer and CodeTalker across all four metrics. This demonstrates the superior capability of our method in generating more precise lip motions and expressions. Furthermore, we also evaluated our method on the TalkingHead-1KH dataset, and the corresponding comparison results are presented in Table~\ref{table:quantitative_th}, reaffirming the superiority of our method.

\subsection{Qualitative Evaluation}

\begin{figure*}[t]
	\centering
	\includegraphics[width=1.0\textwidth]{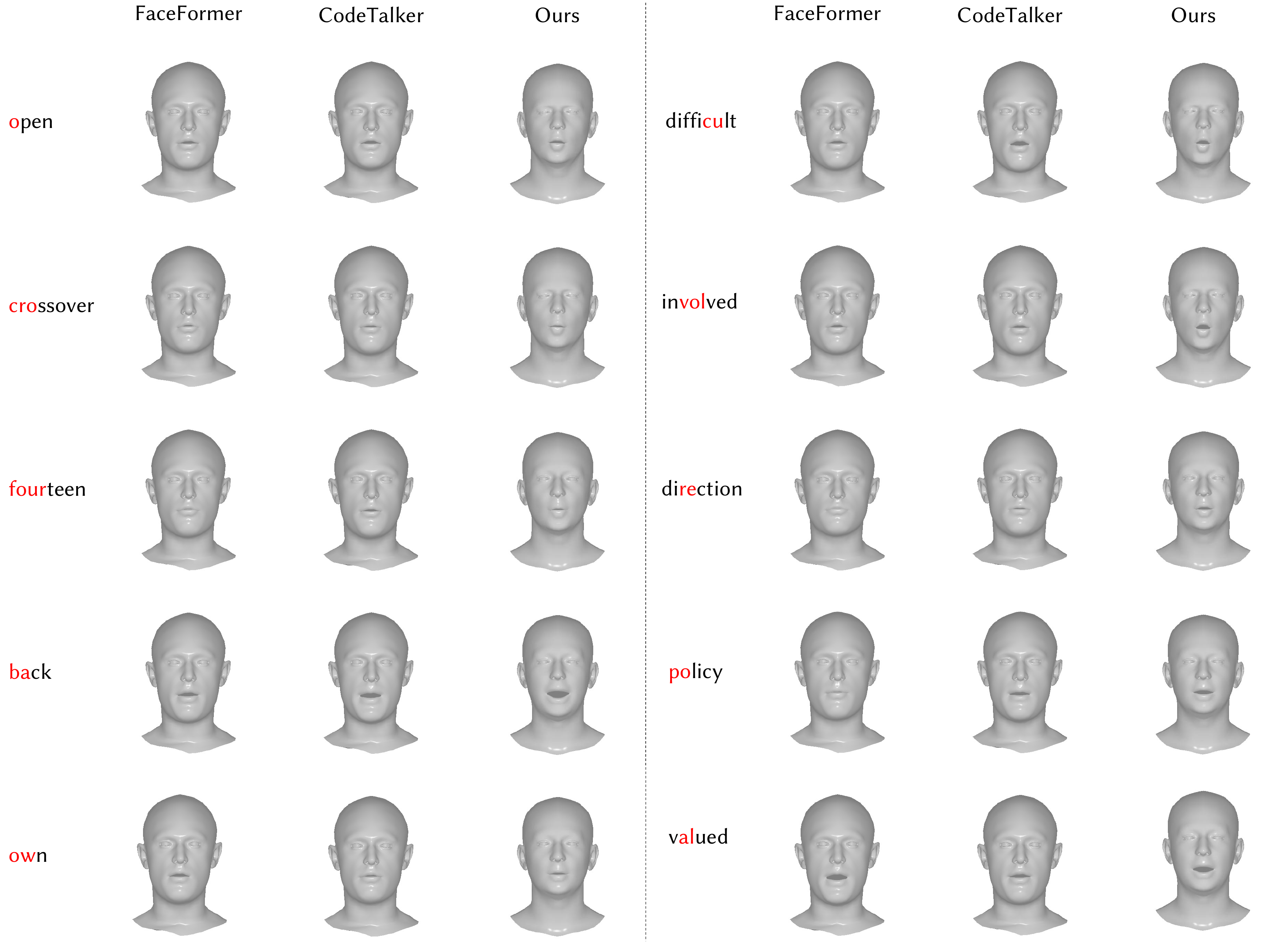}
	\caption{Visual comparisons of sampled facial motions conditioned on different speech parts by different methods.}
	\label{fig:comparison}
  \end{figure*}

We present the qualitative comparison in Fig.~\ref{fig:comparison}. For FaceFormer and CodeTalker, we assign them the same talking style from their style base. As our model possesses distinct talking styles, we randomly select one style for generating the result. It is evident from the comparison that our method excels in producing more precise lip shapes and vivid expressions, enhancing the overall fidelity of the animation.

\subsection{Generalization Capability}

To validate the generalization capability of our model, we conducted tests using head shapes that were not encountered during the training phase. In order to ensure a fair comparison, we also evaluated FaceFormer and CodeTalker using the same audio input and head shapes that were unseen during their respective training. The visual results are presented in Fig.~\ref{fig:change_shape}, showcasing two different cases, with each model being tested using two distinct head shapes. Remarkably, even when applied to unseen head shapes, our model demonstrates the ability to accurately adjust mouth shapes. Conversely, FaceFormer and CodeTalker tend to generate erroneous mouth shapes under similar conditions. This emphasizes the superior generalization capability of our model. More visual comparisons can be viewed in the supplementary video.

\subsection{User Study}

We conducted a user study to assess the lip synchronization and motion realism of our methods. We enlisted 27 participants to take part in the study. Each participant was presented with 10 cases, each comprising three videos: one generated by FaceFormer, another by CodeTalker, and the third by our model. All three videos were driven by the same 10-second audio clip from the TalkingHead-1KH dataset.

\begin{figure*}[t]
	\centering
	\includegraphics[width=0.8\linewidth]{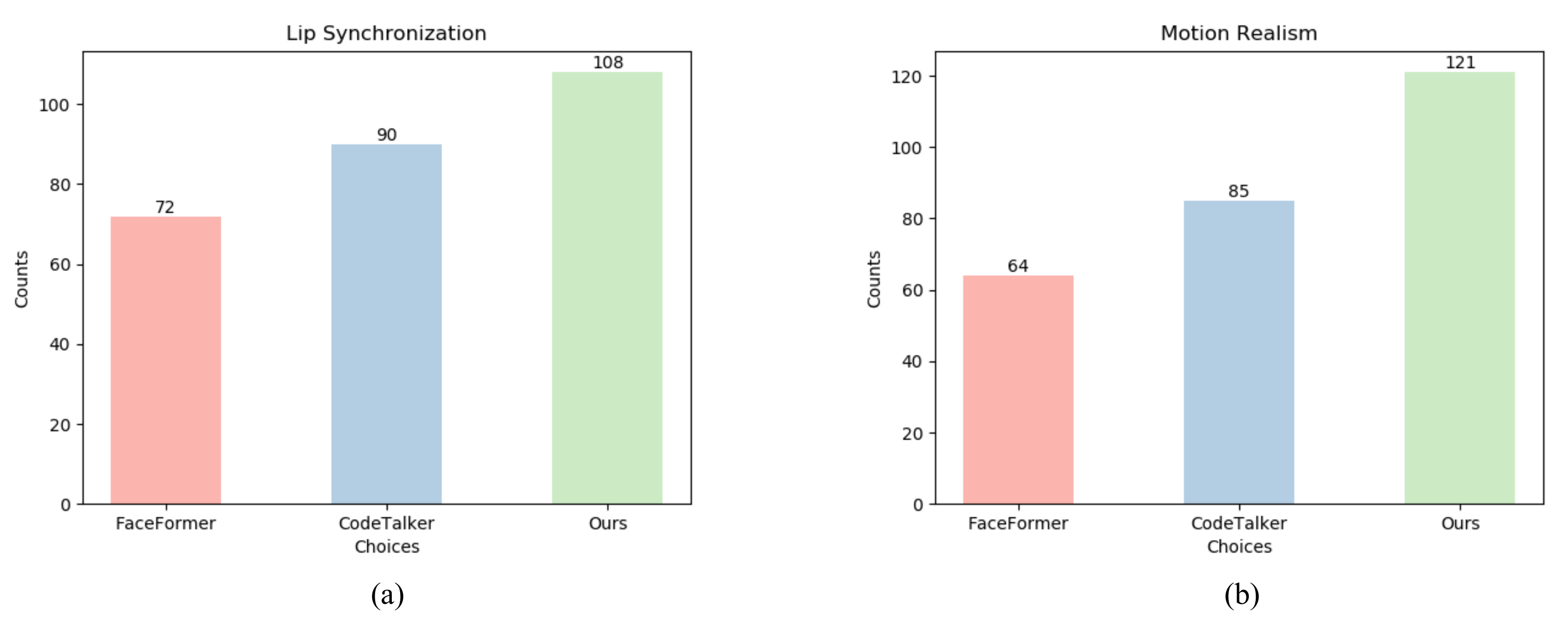}
	\caption{User study results of lip synchronization and motion realism performance across three methods on the TalkingHead-1KH dataset. Participants were asked to select the method with the highest lip synchronization and motion realism.}
	\label{fig:user_sync}
  \end{figure*}

Participants were then asked to select the video that exhibited the highest lip synchronization and the one that displayed the highest motion realism. The results for lip synchronization are presented in Fig.~\ref{fig:user_sync}(a), indicating that our method was predominantly preferred by participants compared to the other two methods. Similarly, Fig.~\ref{fig:user_sync}(b) illustrates the results for motion realism, revealing that our method received the highest preference (121 times) from participants in this aspect.
These results further support the superior performance of our approach.

\subsection{Expression Manipulation}

\begin{figure*}[t]
	\centering
	\includegraphics[width=1.0\textwidth]{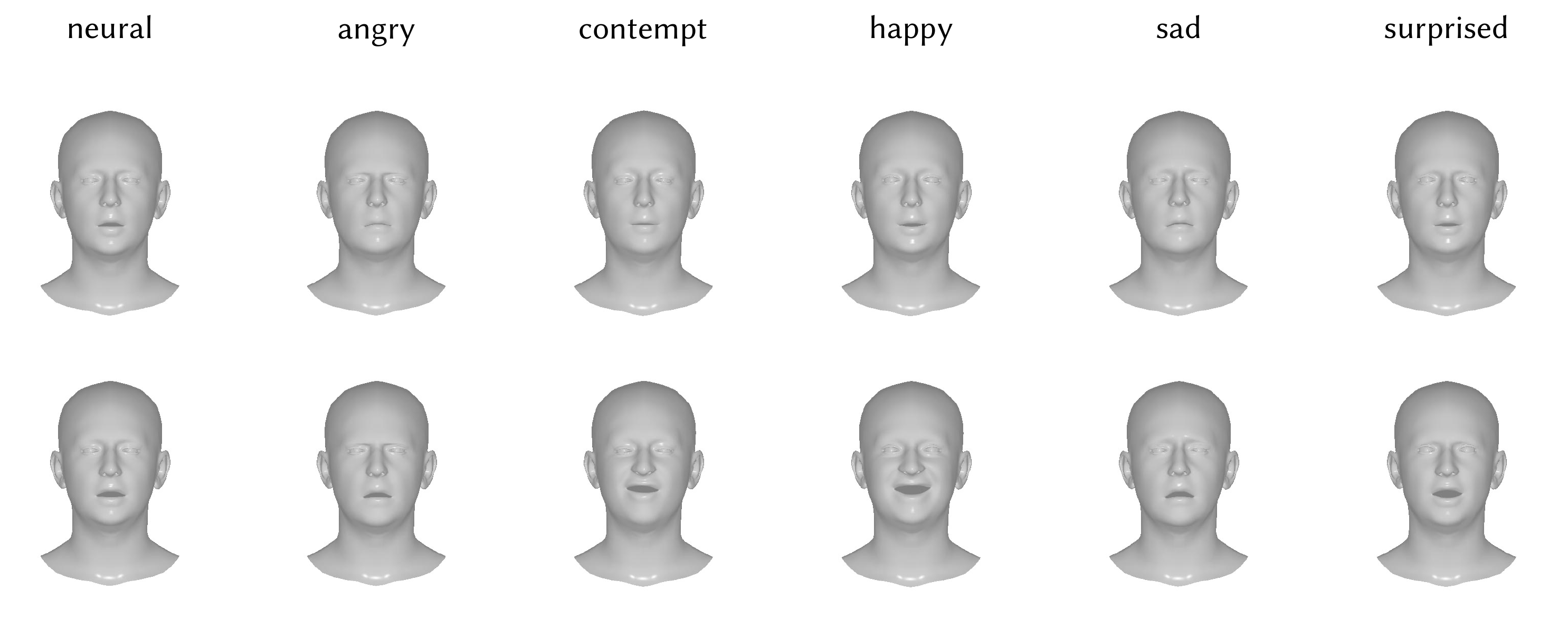}
	\caption{The effectiveness of expression manipulation. By providing different expression labels, one can see that our model can generate the corresponding results.}
	\label{fig:expression}
  \end{figure*}

As introduced in Sec.~\ref{sec:manipulation}, our model is endowed with the capability of expression manipulation. We show some examples in Fig.~\ref{fig:expression}. The first row illustrates examples that the weighting vector $\bm{w}$ is set to $0.4$, while the second row demonstrates cases where it is set to $0.8$.
It is apparent that our model effectively generates results with the intended emotions, such as "angry," "happy," and "sad," while preserving accurate lip shapes. Moreover, a larger value of the weighting vector corresponds to more significant amplitudes in the emotions generated by our model.

\subsection{Ablation Studies}

\begin{table}[t]
	\centering
		\scalebox{1.1}{\begin{tabular}{l|cccc}
			\hline 
			Reg & \cmark & \cmark & \cmark & \cmark \\ 
			MC & \xmark & \cmark & \cmark & \cmark \\ 
			Pho & \cmark & \xmark & \cmark & \cmark \\ 
			Emo & \cmark & \cmark & \xmark & \cmark \\ 
			\hline
			LDE~$\downarrow$ & 0.00094 & 0.00095 & 0.00095 & 0.00094 \\ 
			LVE~$\downarrow$ & 0.00106 & 0.00105 & 0.00106 & 0.00105 \\ 
			EDE~$\downarrow$ & 0.0014 & 0.0013 & 0.0014 & 0.0014 \\ 
			EVE~$\downarrow$ & 0.00054 & 0.00053 & 0.00054 & 0.00053 \\ 
			\hline 
	\end{tabular}}
	\caption{Ablation study on different losses. }
	\label{table:aba_loss}
\end{table}

  \begin{figure*}[t]
	\centering
	\includegraphics[width=1.0\textwidth]{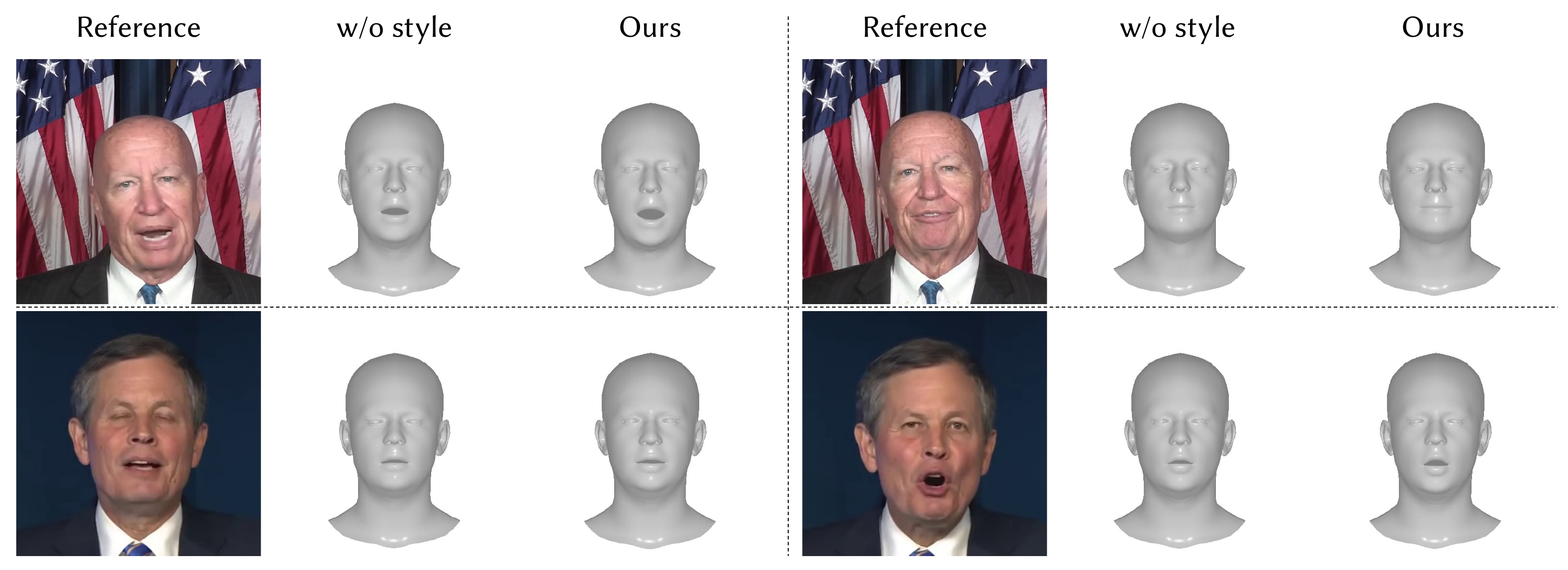}
	\caption{Ablation study on the style code, which will influence the mouth closure and expression amplitudes.}
	\label{fig:woid}
  \end{figure*}

In this section, we conduct ablation studies to investigate different modules of our method, including the effect of the style code and the influence of different loss functions.

\vspace{5pt}
\noindent\textbf{Effect of the Style Code}
The style code is used to control the speaking style of the output video. We study its effect by training a model without the style code. The visual results are shown in Fig.~\ref{fig:woid}. Without the style code, the speaking styles of generated videos driven by different audios tend to be similar. 
Conversely, when the style code is utilized, different videos can reflect different speaking styles. For instance, in the top left corner of the figure, the first example demonstrates more pronounced mouth-opening amplitudes, validating the influence of the style code in capturing diverse speaking styles.

\vspace{5pt}
\noindent\textbf{Influence of different losses}
We further study the influence of different losses mentioned in Sec.~\ref{sec:loss}. Table.~\ref{table:aba_loss}. It can be observed that with the full objective function, our model achieves the best performance.

\begin{figure}[t]
  \centering
  \includegraphics[width=0.9\linewidth]{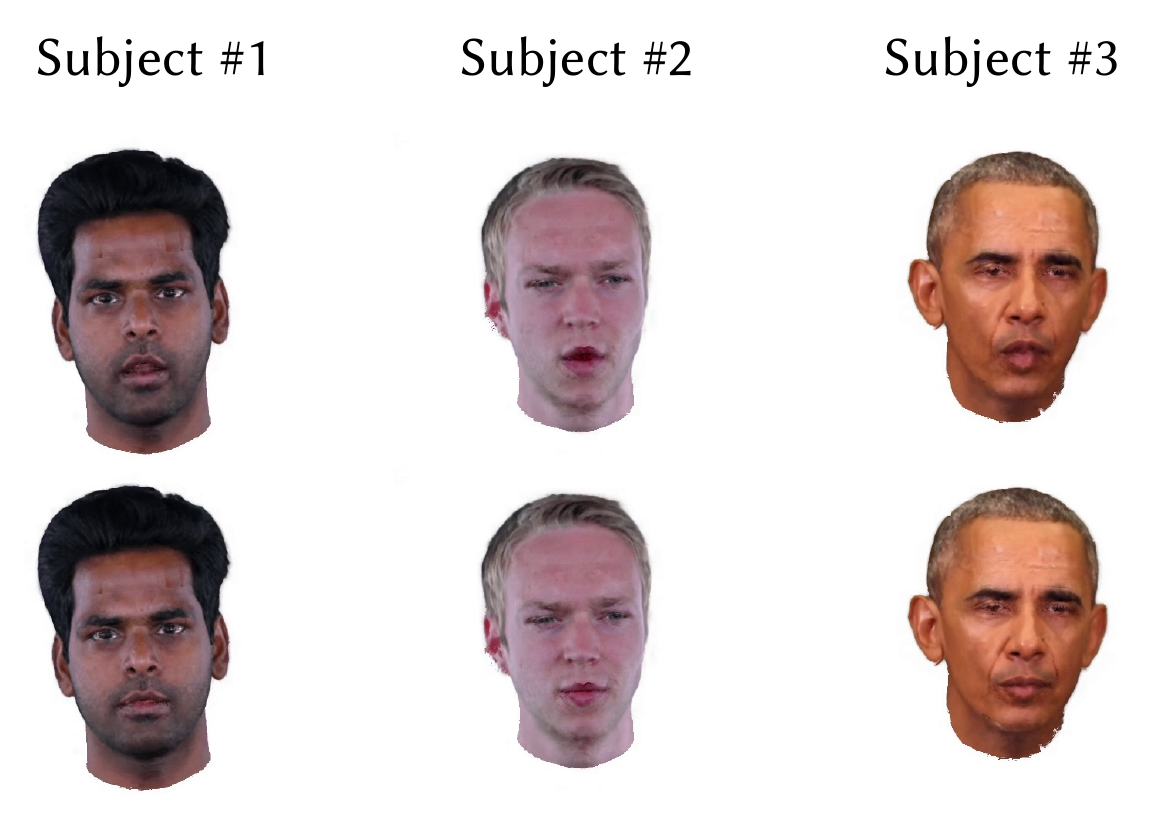}
  \caption{Our method can be combined with INSTA~\cite{zielonka2022instant} to synthesize photo-realistic avatars.}
  \label{fig:application}
\end{figure}

\subsection{More Applications}

Aside from generating 3D talking meshes, our model can also provide the facial motion information for synthesizing photo-realistic avatars. In Fig.~\ref{fig:application}, we show some examples generated by combining our method with INSTA~\cite{zielonka2022instant}. The results clearly demonstrate the accuracy and robustness of our model in producing lifelike facial motions, even when dealing with diverse subjects.

\section{Conclusion and Limitations}

In this paper, we mainly focus on addressing the audio-driven 3D facial animation problem. Unlike existing methods that heavily rely on training their models using limited public 3D datasets, which offer only a restricted number of audio-3D scan pairs, we propose a novel method that leverages in-the-wild 2D talking-head videos for training our 3D facial animation model. The availability of these 2D videos is abundant, and their diverse range of identities and facial motions can greatly enhance the performance of our model. Consequently, our model exhibits robust generalization capabilities and consistently produces high-fidelity lip synchronization. Moreover, our model can also proficiently captures the unique identities and speaking styles of various individuals.

\vspace{5pt}
\noindent\textbf{Limitations}
However, our model does have a few limitations that are worth mentioning. Firstly, the simplicity of the audio encoder we employed can make it sensitive to noise in some instances. To address this issue, one potential solution is to utilize pre-trained speech models, such as wav2vec 2.0~\cite{baevski2020wav2vec}, as an alternative audio encoder. This could enhance the model's ability to handle noisy audio inputs.
Secondly, during expression manipulation, the value of the weighting vector is manually set, resulting in fixed emotion amplitudes. To introduce more dynamic and vivid emotion changes during speaking, a potential future direction is to employ a small network to learn the weighting vector. This would enable the model to dynamically adjust and modulate emotions in a more nuanced manner.
By addressing these limitations, we can further improve the robustness and flexibility of our model in handling different audio conditions and achieving more expressive and natural results.

{\small
	\bibliographystyle{ieee_fullname}
	\bibliography{egbib}
}

\clearpage
\newpage


	

\end{document}